\documentclass{optica-article}

\journal{opticajournal} 

\articletype{Research Article}

\usepackage{subcaption}
\usepackage{booktabs}

\begin{document}

\title{Retinal Layer Segmentation in OCT Images With 2.5D Cross-slice Feature Fusion Module for Glaucoma Assessment}

\author{Hyunwoo Kim,\authormark{1,$\dagger$} Heesuk Kim,\authormark{2,$\dagger$} Chaewon Lee,\authormark{3} Wungrak Choi,\authormark{2,*} and Jae-Sang Hyun\authormark{1,**}}


\address{\authormark{1}Department of Mechanical Engineering, Yonsei University, Seoul, 03722, South Korea\\\authormark{2}Institute of Vision Research, Department of Ophthalmology, Severance Hospital, Yonsei University, Seoul, 03722, South Korea\\\authormark{3}Yonsei University College of Medicine, Seoul, 03722, South Korea\\\textup{$\dagger$These authors contributed equally.}}

\email{corresponding authors: \authormark{*}wungrakchoi@yuhs.ac, \authormark{**}hyun.jaesang@yonsei.ac.kr} 


\begin{abstract*}
For accurate glaucoma diagnosis and monitoring, reliable retinal layer segmentation in OCT images is essential.
However, existing 2D segmentation methods often suffer from slice-to-slice inconsistencies due to the lack of contextual information across adjacent B-scans.
3D segmentation methods are better for capturing slice-to-slice context, but they require expensive computational resources.
To address these limitations, we propose a 2.5D segmentation framework that incorporates a novel cross-slice feature fusion (CFF) module into a U-Net-like architecture.
The CFF module fuses inter-slice features to effectively capture contextual information, enabling consistent boundary detection across slices and improved robustness in noisy regions.
The framework was validated on both a clinical dataset and the publicly available DUKE DME dataset.
Compared to other segmentation methods without the CFF module, the proposed method achieved an 8.56\% reduction in mean absolute distance and a 13.92\% reduction in root mean square error, demonstrating improved segmentation accuracy and robustness.
Overall, the proposed 2.5D framework balances contextual awareness and computational efficiency, enabling anatomically reliable retinal layer delineation for automated glaucoma evaluation and potential clinical applications.
\end{abstract*}

\section{Introduction}\label{sec:introduction}

Glaucoma is one of the primary causes of irreversible blindness worldwide, with projections estimating that 111.8 million individuals will be affected by 2040. Since 25\%-35\% of retinal ganglion cells (RGCs) may be lost before visual field defects become detectable, early detection is crucial for preventing irreversible vision loss.
Optical coherence tomography (OCT) now performs a central role in early diagnosis by providing high-resolution visualizations of retinal structures~\cite{harwerth1999ganglion,kerrigan2000number,tham2014global}.
Macular OCT is particularly valuable because it can detect early damage that standard perimetry might miss, as approximately 50\% of RGCs are concentrated in the macula~\cite{hood2013glaucomatous,weinreb2014pathophysiology}.
Moreover, segmentation of the retinal layer in OCT images allows clinicians to quantify clinically meaningful biomarkers such as retinal layer thickness, making OCT even more effective for detecting early structural damage.

Despite its diagnostic value, OCT analysis still faces practical limitations.
Widely used segmentation algorithms often fail on images of eyes with pathologies such as high myopia or epiretinal membranes, leading to errors in approximately 8.5\% of scans\cite{chiu2010automatic,dufour2012graph,mansberger2017automated,yang2025diagnosis}.
These discrepancies, exacerbated by device and protocol variability~\cite{kim2025establishment,tan2012comparison}, necessitate manual correction by clinicians.
In busy clinical settings, correcting these segmentation errors manually is labor-intensive and introduces inter-observer variability, creating a barrier to high-throughput glaucoma screening workflows.
Given that layer-specific parameters such as the ganglion cell-inner plexiform layer are critical for diagnosis~\cite{mahmoudinezhad2023comparison}, especially when the retinal nerve fiber layer (RNFL) reaches a measurement floor in advanced disease~\cite{hood2013glaucomatous,weinreb2014pathophysiology}, there is an urgent need for robust, automated segmentation methods~\cite{shi2024artificial}.

Recently, deep learning approaches have been actively explored and demonstrated strong performance in accurate retinal layer segmentation.
However, the conventional 2D-based segmentation methods treat each B-scan as an isolated entity, processing it independently without knowledge of the surrounding anatomy. This is analogous to reading a single sentence out of context; if a specific B-scan is degraded by speckle noise or vessel shadows, a 2D model lacks the context to distinguish true anatomical boundaries from artifacts, leading to spike-like segmentation errors.
By contrast, 3D-based methods use the entire volumetric data to capture this spatial context, but they require immense computational power and memory~\cite{li2006optimal,garvin2009automated,sleman2021novel}.
This often necessitates downsampling the image resolution to make processing feasible, which can cause the fine structural details necessary for precise early glaucoma diagnosis to be lost.

Recently, 2.5D segmentation has emerged as third option that balances accuracy with clinical feasibility~\cite{hung2022cat,hung2024csam,kumar2024flexible}.
The core concept of 2.5D segmentation is to mimic the human expert’s review process.
When a clinician encounters an ambiguous layer boundary due to noise in a specific B-scan, they typically scroll to adjacent anterior and posterior B-scans to verify anatomical continuity.
Similarly, a 2.5D framework inputs a stack of neighboring slices to predict the boundaries of the central slice. By incorporating these cross-slice contextual cues, the model can correct a noisy region in one slice by referencing clear data from its neighbors, ensuring structural consistency without the prohibitively high computational cost of a full 3D model.

Based on this idea, we propose a 2.5D segmentation framework that leverages a novel cross-slice feature fusion (CFF) module to effectively integrate contextual information from adjacent B-scans.
Our CFF module is highly adaptable, as it replaces the traditional skip connections in existing encoder-decoder architectures, which are widely used in segmentation networks.
In this work, we aim to detect five key boundaries that are critical for glaucoma diagnosis: the internal limiting membrane (ILM) and the surfaces between the RNFL and ganglion cell layer (RNFL-GCL), inner plexiform layer and inner nuclear layer (IPL-INL), outer plexiform layer and outer nuclear layer (OPL-ONL), and outer nuclear layer and inner segment (ONL-IS).

We validated our proposed method on both real clinical data and publicly available datasets.
Moreover, we quantified the improvements using the differences in boundary position with respect to established baselines and showed that more accurate delineations can enhance the reliability and practical value of OCT across diverse patient populations.

\section{Related Works}\label{sec:related_works}

With the advancement of image processing technologies, various automated retinal layer segmentation methods have been developed. 
Earlier approaches primarily relied on traditional image processing techniques, while more recent studies have leveraged the progress of deep learning to propose data-driven segmentation methods. 
This section introduces various existing methods, including traditional methods, deep learning methods, and 2.5D segmentation methods.

\subsection{Traditional methods}
Traditional methods focused on detecting retinal layer boundaries in a column-wise manner. 
Some studies attempted to segment layers based on pixel intensity variations in OCT images. 
Ishikawa et al.~\cite{ishikawa2005macular} analyzed the intensity profile along each image column to identify retinal boundaries. 
Fabritius et al.~\cite{fabritius2009automated} proposed a method for detecting the two boundaries with the highest contrast, ILM and RPE, by exploiting intensity gradients. 
Shahidi et al.~\cite{shahidi2005quantitative} employed curve fitting to enforce smoothness constraints, assuming that the retinal boundaries form continuous curves.
These methods are computationally efficient, but have limitations in handling noisy or low contrast regions, which are common in OCT images.

Another category of approaches is based on active contour models.
These methods formulate segmentation as an energy-minimization problem under shape priors, where the retinal boundaries are assumed to be smooth and continuous. 
Yazdanpanah et al.~\cite{yazdanpanah2009intra} applied the Chan-Vese active contour model combined with an adaptive weighting strategy to extract multiple retinal boundaries. 
Mishra et al.~\cite{mishra2009intra} extended the classical Kass Snakes model~\cite{kass1988snakes} using a two-step kernel-based optimization scheme. 
Rossant et al.~\cite{rossant2015parallel} introduced the Parallel Double Snakes model, which simultaneously evolves two adjacent contours under a soft parallelism constraint. 
These methods can handle noisy or low-contrast regions better than intensity variation-based methods, but their performance strongly depends on the definition of the shape priors, which may limit robustness across variable anatomical structures.

To overcome these limitations, graph-based methods have been proposed. 
These approaches construct a graph from the OCT image and detect optimal layer boundaries through energy minimization, offering greater robustness to noise. 
Chiu et al.~\cite{chiu2010automatic} estimated graph weights from intensity gradients and applied dynamic programming to solve a shortest-path problem for boundary detection. 
Dufour et al.~\cite{dufour2012graph} improved this approach by incorporating hard and soft constraints derived from a learned statistical model. 
Kaba et al.~\cite{kaba2015retina} proposed a multi-regional kernel graph-cut algorithm based on the max-flow formulation to segment retinal layers.
While these methods are more robust to noise and can handle complex anatomical structures better than earlier methods, they are sensitive to initial node selection and often require careful parameter tuning.

\subsection{Deep learning methods}
Recently, convolutional neural networks (CNNs) are widely used in various image processing tasks, leading to the development of numerous deep learning-based retinal layer segmentation methods.
Some studies have combined deep learning with graph-based methods.
Fang et al.~\cite{fang2017automatic} used CNN for feature extraction, followed by the generation of probability maps, and then applied graph search methods to detect the final retinal layer boundaries.
Hu et al.~\cite{hu2019automatic} proposed a multiscale convolutional neural network to extract multiscale features of retinal layer boundaries and produce probability maps of the retinal layer boundaries. 
Raja et al.~\cite{raja2020extraction} employed the VGG-16 architecture~\cite{simonyan2014very} for feature extraction, followed by the generation of probability maps.

The emergence of fully convolutional neural networks (FCNs)~\cite{long2015fully} and U-Net~\cite{ronneberger2015u}, which have shown remarkable performance in image segmentation tasks, has inspired attempts to apply end-to-end frameworks to retinal layer segmentation. 
Li et al.~\cite{li2020deepretina} proposed DeepRetina, an end-to-end FCN framework, and Roy et al.~\cite{roy2017relaynet} proposed ReLayNet, an end-to-end U-Net framework. 
Particularly, the encoder-decoder structure of U-Net has proven highly effective for image segmentation, leading to the development of numerous frameworks based on this architecture. 
Wang et al.~\cite{wang2021boundary} proposed a boundary-aware U-Net, a dual-task framework with low-level outputs for boundary detection and high-level outputs for layer segmentation. 
He et al.~\cite{he2021structured} introduced a method with two output branches and distinct loss functions for each output, enabling simultaneous pixel-wise segmentation and column-wise boundary detection. 
They applied topological constraints to the loss function for boundary detection, achieving smooth, continuous, and topologically correct boundaries. 
These deep learning-based methods have enabled more robust and accurate segmentation, but they still face limitations in ensuring contextual information across multiple consecutive OCT images.

\subsection{2.5D segmentation methods}
2.5D segmentation methods utilize multiple consecutive images and 2D CNNs, offering lower computational cost compared to 3D methods while better preserving contextual information across OCT images than 2D methods. 
Some studies~\cite{li2021acenet, wang2019benchmark, duan2019automatic, xie2022automated} concatenate multiple consecutive grayscale images along the channel dimension to form a single multi-channel image. 
This combined image is then fed into a 2D segmentation network to obtain the segmentation result for the middle slice image. 
However, concatenating multiple images into a single multi-channel image complicates the segmentation of individual slices and may reduce segmentation accuracy.

To address these limitations, other studies~\cite{yu2018recurrent, yang2018towards} employed recurrent neural networks (RNNs), which have shown outstanding performance in natural language processing tasks where contextual information is essential. 
In these methods, each slice image was treated as a temporal sequence and processed using an RNN architecture. 
However, these approaches are computationally expensive.

To reduce computational cost, methods leveraging attention mechanisms have been proposed~\cite{hung2022cat, hung2024csam, kumar2024flexible}. 
These methods replace the skip connection in U-Net-based encoder-decoder architectures with cross-slice attention modules to incorporate contextual information from neighboring slices. 
CAT-Net~\cite{hung2022cat} and CSAM~\cite{hung2024csam} models utilized cross-slice attention and transformer blocks to learn correlations between consecutive images. 
CSA-Net~\cite{kumar2024flexible} model further added an in-slice attention module to understand correlations among pixels within the center slice. 
These studies highlight the growing interest in 2.5D segmentation models and their potential to address the trade-offs between 2D and 3D approaches.

\section{Methods}\label{sec:methods}

\begin{figure}[tbp]
    \centering
    \includegraphics[width=\linewidth]{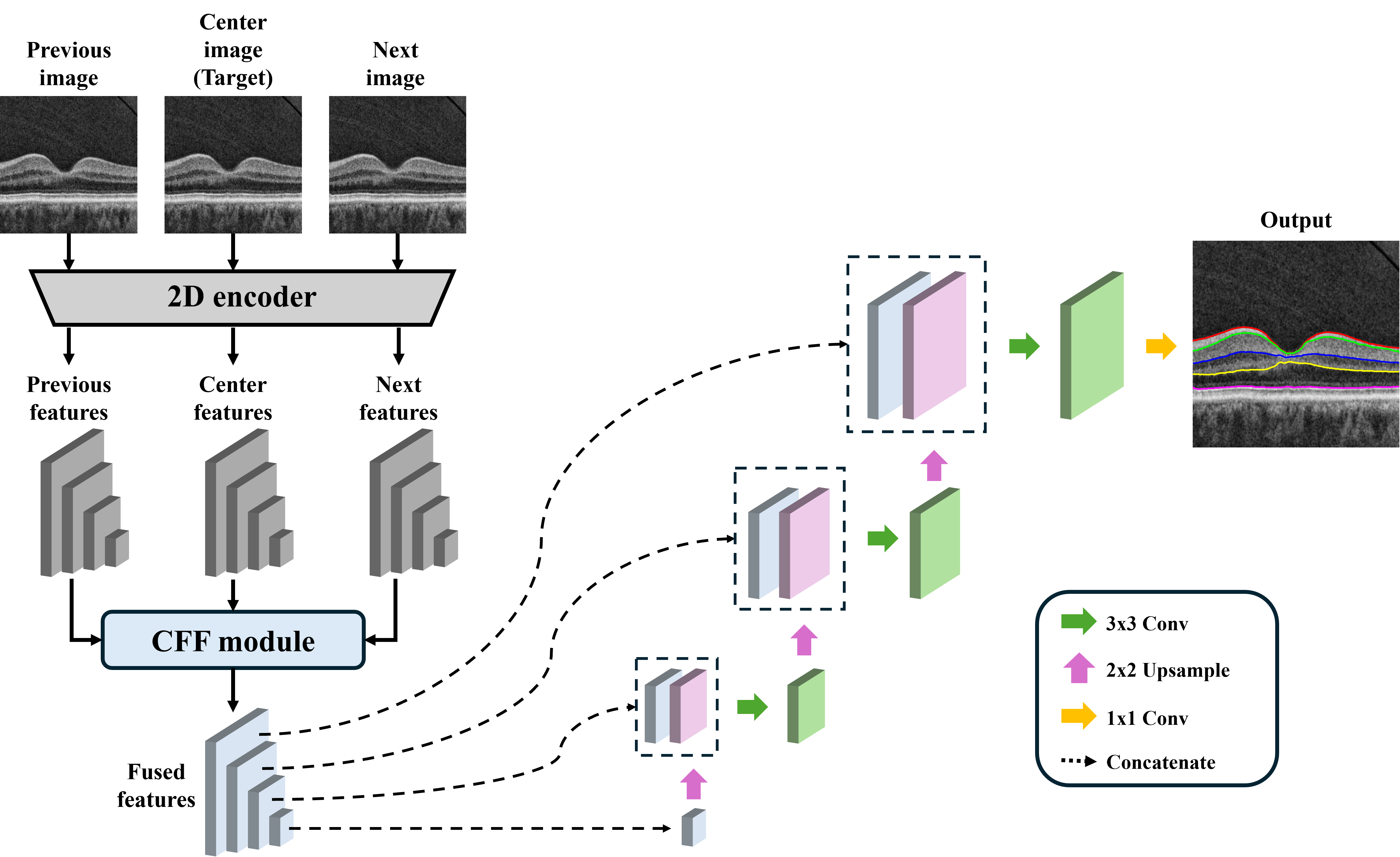}
    \caption{Overview of the proposed method.}
    \label{fig:overview}
\end{figure}

\subsection{Overview}
We propose a 2.5D Cross-slice Feature Fusion (CFF) module designed to capture inter-slice contextual information for retinal layer segmentation.
Our CFF module-based 2.5D segmentation network consists of 2D encoders, 2D decoders, and CFF modules placed between the encoders and decoders.
The network takes \(n\) consecutive slices (\(n = 3, 5, 7, \dots\)) as input and predicts the segmentation map of the center slice.
The \(n\) images independently pass through each level of the encoder, generating \(n\) feature maps at each level.
These feature maps are fused into a single feature map through the CFF module.
The fused feature map is then passed to the corresponding level of the decoder.
Since the skip connections in the original U-Net are simply replaced with CFF modules, our approach can be applied to any U-Net-like architecture.
Fig.~\ref{fig:overview} shows an overview of our proposed method.

\subsection{Pre-processing}\label{sec:preprocessing}
The OCT images were pre-processed before being fed into the network.
First, the images were flattened to reduce variability caused by the natural curvature of the retina.
This was crucial for accurate segmentation, as the curvature could lead to misalignment of the retinal layers across different images.
The flattening process involved detecting the RPE layer and aligning it to the central row of the image.
This ensured that the RPE layer was consistently positioned, allowing for better comparison and segmentation across images.
The RPE layer has the highest intensity in OCT images, while the ONL layer directly above it has the lowest intensity among all layers.
Based on this prior knowledge, we identified the point with the largest intensity gradient in each column of the image.
If the point position differed by more than 60 pixels from its neighboring columns, it was considered an outlier and removed.
The final RPE layer was obtained by fitting the remaining points to a second-order polynomial, which provided a smooth and continuous curve.
The image columns were vertically shifted so that the detected RPE layer aligned with the central row of the image.

\begin{figure}[htbp]
    \centering
    \includegraphics[width=\linewidth]{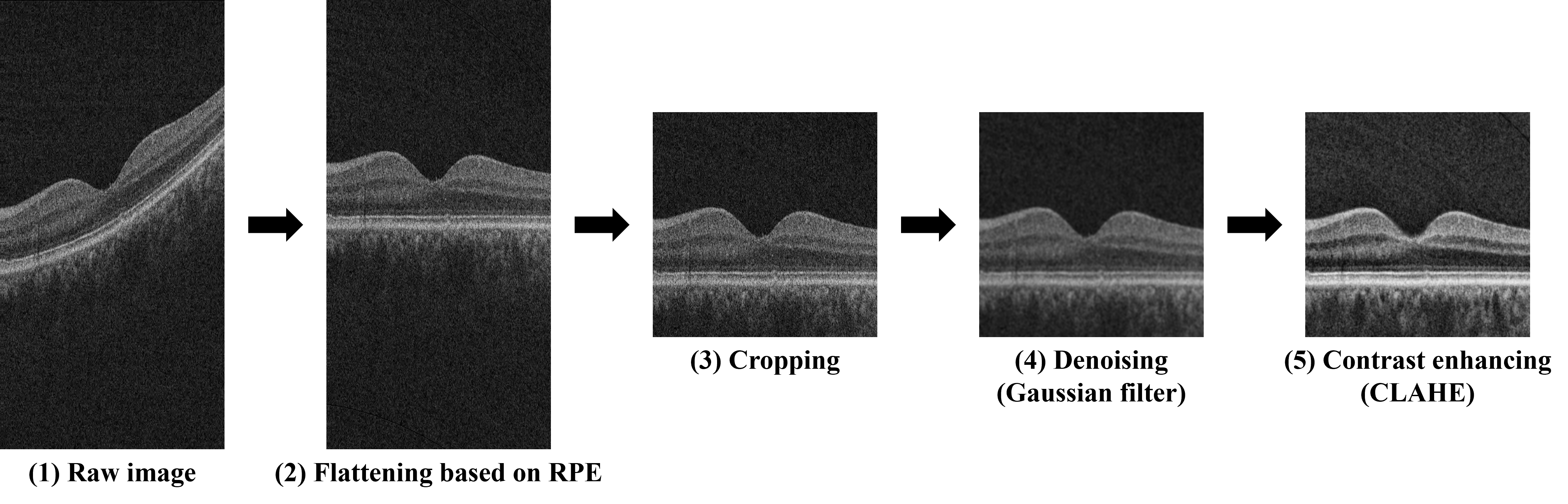}
    \caption{Pre-processing steps for OCT images.}
    \label{fig:preprocessing}
\end{figure}

After image flattening, we sequentially performed cropping, noise reduction, and contrast enhancement.
The top and bottom ends of the image correspond to the vitreous and choroid regions, which are background areas without retinal layers.
To focus on the retinal layers, we cropped the image to retain the region spanning from 1/8 to 5/8 of the total height.
Next, we applied a Gaussian filter to smooth the image and reduce noise.
Finally, we enhanced the contrast of the image using contrast-limited adaptive histogram equalization (CLAHE)~\cite{van2014scikit} which improves the visibility of retinal layers by locally enhancing the contrast in different regions of the image.
Fig.~\ref{fig:preprocessing} illustrates the overall image pre-processing steps and provides examples of the results.

\subsection{Cross-slice feature fusion (CFF) module}
The CFF module is designed to fuse feature maps extracted from multiple consecutive slices, enabling the network to capture inter-slice contextual information.
It performs feature aggregation using a pixel-wise weighted sum mechanism, where spatially adaptive weights are learned to reflect the relative importance of each slice at every spatial location.
These weights are predicted by CNN layers that take the feature maps of each slice as input.
This approach is particularly effective in regions affected by speckle noise or vessel-induced artifacts, where certain slices may contain degraded or ambiguous features.
In such cases, the CFF module can assign lower weights to unreliable slices and higher weights to more informative slices, thus enhancing the overall feature representation for accurate segmentation.

Given \(n\) consecutive OCT slices, each of the \(n\) slices is independently passed through a shared-weight encoder to produce a feature map \(F_i \in \mathbb{R}^{h \times w \times c}\), where \(i = 1, \dots, n\). 
Here, \(h\), \(w\), and \(c\) denote the height, the width, and the number of channels of the feature map, respectively. 
The resulting \(n\) feature maps \(\{F_1, F_2, \dots, F_n\}\) are concatenated along the channel dimension to form a stacked feature map \(\mathbf{F} \in \mathbb{R}^{h \times w \times nc}\), which aggregates all contextual information from the adjacent slices.

To estimate the relative importance of each input slice, the stacked feature map \(\mathbf{F}\) is passed through a \(1 \times 1\) convolutional layer, producing a weight volume \(\mathbf{W} \in \mathbb{R}^{h \times w \times n}\). 
Each channel of \(\mathbf{W}\), denoted as \(W_i \in \mathbb{R}^{h \times w}\), corresponds to a spatial weight map for the \(i\)-th slice.
Since these weight maps are intended to assign slice-wise importance at each spatial location, a softmax operation is applied across the slice dimension to normalize the weights.

\begin{equation}
W_i(x, y) = \frac{\exp(\mathbf{W}(x, y, i))}{\sum_{j=1}^{n} \exp(\mathbf{W}(x, y, j))}, \quad \forall i \in \{1, \dots, n\}
\end{equation}
The normalized weights ensure that the sum of weights across slices at each pixel location \((x, y)\) equals 1, preventing the resulting weighted sum from exceeding the range of the input values.

The final fused feature map \(F_{\text{fused}} \in \mathbb{R}^{h \times w \times c}\) is computed through a pixel-wise weighted sum of the feature maps \(\{F_1, F_2, \dots, F_n\}\) with a residual connection as follows

\begin{figure}[tbp]
    \centering
    \includegraphics[width=\linewidth]{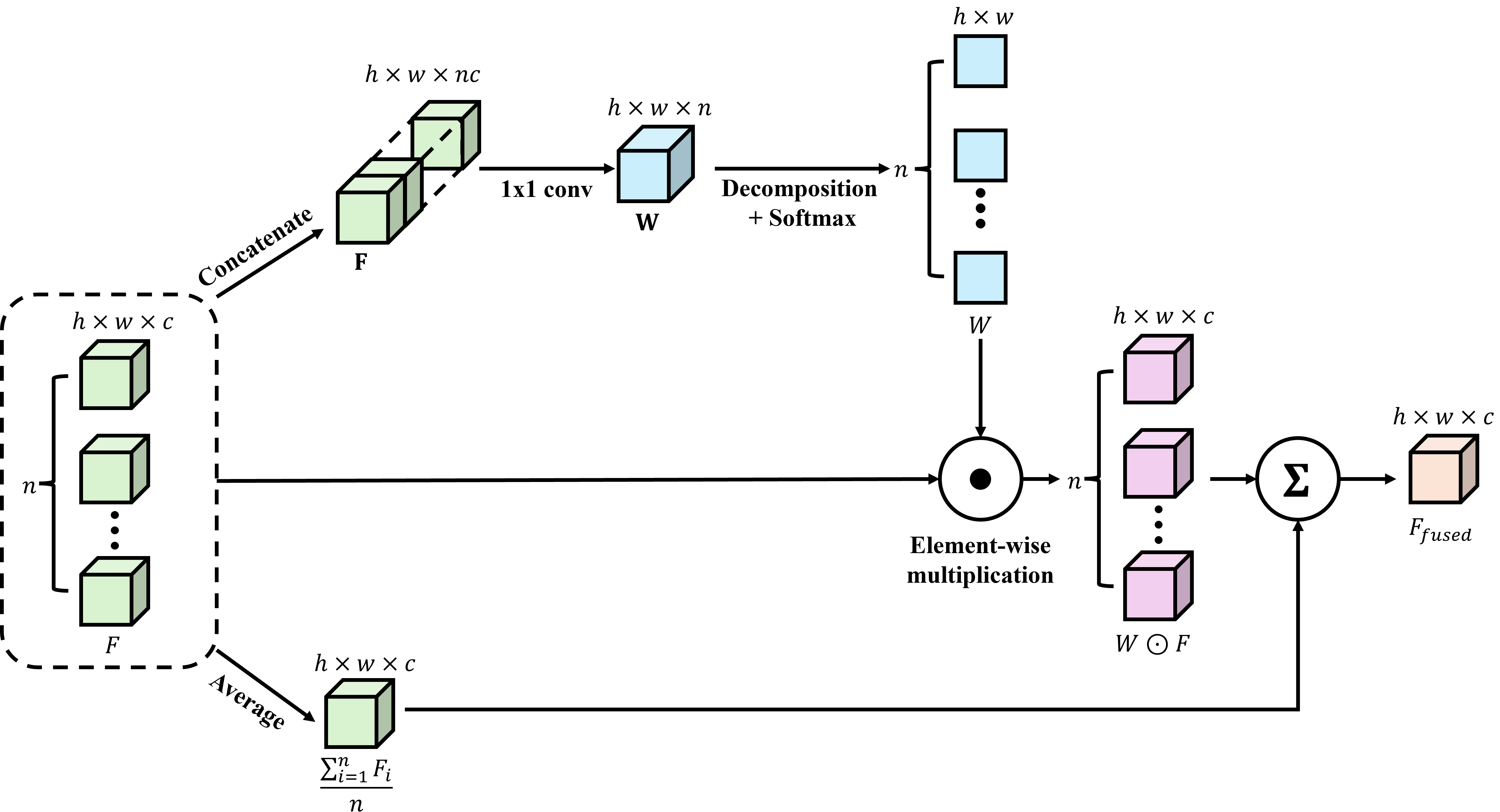}
    \caption{Overview of cross-slice feature fusion (CFF) module with input \(F\) and output \(F_{\text{fused}}\).}
    \label{fig:cff_module}
\end{figure}

\begin{equation}
F_{\text{fused}} = \sum_{i=1}^{n} W_i \odot F_i + \frac{1}{n} \sum_{i=1}^{n} F_i
\end{equation}
where \(\odot\) denotes element-wise multiplication. 
The residual connection stabilizes the fusion process by allowing the network to retain the baseline representation while incorporating the weighted features.
Finally, the fused feature map \(F_{\text{fused}}\) is passed to the corresponding decoder stage.
Fig.~\ref{fig:cff_module} illustrates the mechanism of the CFF module with \(n\) consecutive OCT slices.

\subsection{Network architecture}
The proposed CFF module can be integrated into various U-Net-like architectures.
In this work, we adopted the Fully Convolutional Residual Network (FCRN)~\cite{he2021structured} as the backbone for our segmentation network.
The FCRN is built upon a residual U-Net backbone and contains two output branches: one for pixel-wise segmentation (conv-m) and another for column-wise boundary detection (conv-s).
In our implementation, we replaced all skip connections in the residual U-Net encoder-decoder backbone with the proposed CFF module.
Specifically, given \(n\) consecutive OCT slices, each slice is independently passed through a shared encoder, and the resulting feature maps at each scale are fused using the CFF module to form a single, context-enriched feature map.
This fused feature map is then passed to the corresponding decoder level.
Finally, the network predicts the segmentation map and layer boundaries for the center slice among the \(n\) input slices.

The encoder consists of two sequential blocks of 3\(\times\)3 convolution, batch normalization, and PReLU activation, followed by a residual connection.
For downsampling and upsampling, 2\(\times\)2 max-pooling and 2\(\times\)2 bilinear interpolation are used, respectively.
At the end of the decoder, the final feature map is separately passed to conv-m and conv-s branches.
Conv-m branch predicts pixel-wise probability maps for each retinal layer using a channel-wise softmax and conv-s branch predicts surface positions using a column-wise soft-argmax and a topology guarantee module.


The loss function for training is defined as a weighted combination of three loss components.
The first component, \(L_{\text{maskCE}}\), is the cross-entropy loss applied to the output of the conv-m branch, which performs pixel-wise segmentation of retinal layers as follows
\begin{equation}
L_{\text{maskCE}} = -\frac{1}{HW} \sum_{u=1}^{W} \sum_{v=1}^{H}\sum_{c=1}^{C} y_{u,v,c} \log(\hat{y}_{u,v,c})
\end{equation}
where \(W\) and \(H\) is the width and the height of the image respectively, \(C\) is the number of classes, \(y_{u,v,c}\) is the ground truth label for class \(c\) at pixel \((u,v)\), and \(\hat{y}_{u,v,c}\) is the predicted probability for class \(c\) at pixel \((u,v)\).

The second component, \(L_{\text{lineCE}}\), is the cross-entropy loss computed on the output of the conv-s branch, which estimates the vertical positions of anatomical boundaries using column-wise predictions as follows
\begin{equation}
L_{\text{lineCE}} = -\frac{1}{W} \sum_{u=1}^{W} \sum_{v=1}^{H} z_{u,v} \log(\hat{z}_{u,v})
\end{equation}
where \(z_{u,v}\) is the ground truth label for the boundary position at column \(u\) and row \(v\), and \(\hat{z}_{u,v}\) is the predicted probability for the boundary at the same location.

The third component, \(L_{\text{lineL1}}\), is a smooth L1 loss applied to the final structured surface output, encouraging smooth and topologically consistent boundary surfaces as follows
\begin{equation}
L_{\text{lineL1}} = \frac{1}{W} \sum_{u=1}^{W} \ell(s_u, \hat{s}_u)
\end{equation}
where \(s_u\) is the ground truth surface position at column \(u\), \(\hat{s}_u\) is the predicted surface position at the same column, and \(\ell(s_u, \hat{s}_u)\) is defined as follows.
\begin{equation}
\ell(s_u, \hat{s}_u) =
\begin{cases} 
0.5 \cdot (s_u - \hat{s}_u)^2, & \text{if } |s_u - \hat{s}_u| < 1, \\
|s_u - \hat{s}_u| - 0.5, & \text{otherwise}.
\end{cases}
\end{equation}
This loss function combines the benefits of L1 and L2 losses, being less sensitive to outliers while maintaining stability for small deviations.

These three losses are linearly combined to form the total loss.

\begin{equation}
L_{\text{total}} = \lambda_1 L_{\text{maskCE}} + \lambda_2 L_{\text{lineCE}} + \lambda_3 L_{\text{lineL1}}
\end{equation}
Here, \(\lambda_1\), \(\lambda_2\), and \(\lambda_3\) are weighting coefficients that control the contribution of each term during training.

\section{Experiments}\label{sec:experiments}

\begin{table}[t]
\centering
\caption{Distribution of dataset cases across training, validation, and test sets.}
\resizebox{0.74\textwidth}{!}{%
\begin{tabular}{lcccc}
\toprule
                         & Normal & Glaucoma & Other disease & Total \\
\midrule
Train dataset            & 4      & 12       & 4             & 20    \\
Validation dataset       & 0      & 2        & 0             & 2     \\
Test dataset             & 2      & 6        & 2             & 10    \\
\midrule
Total                   & 6      & 20       & 6             & 32    \\
\bottomrule
\end{tabular}%
}
\label{tab:dataset_distribution}
\end{table}

\begin{figure}[t]
    \centering
    \includegraphics[width=\linewidth]{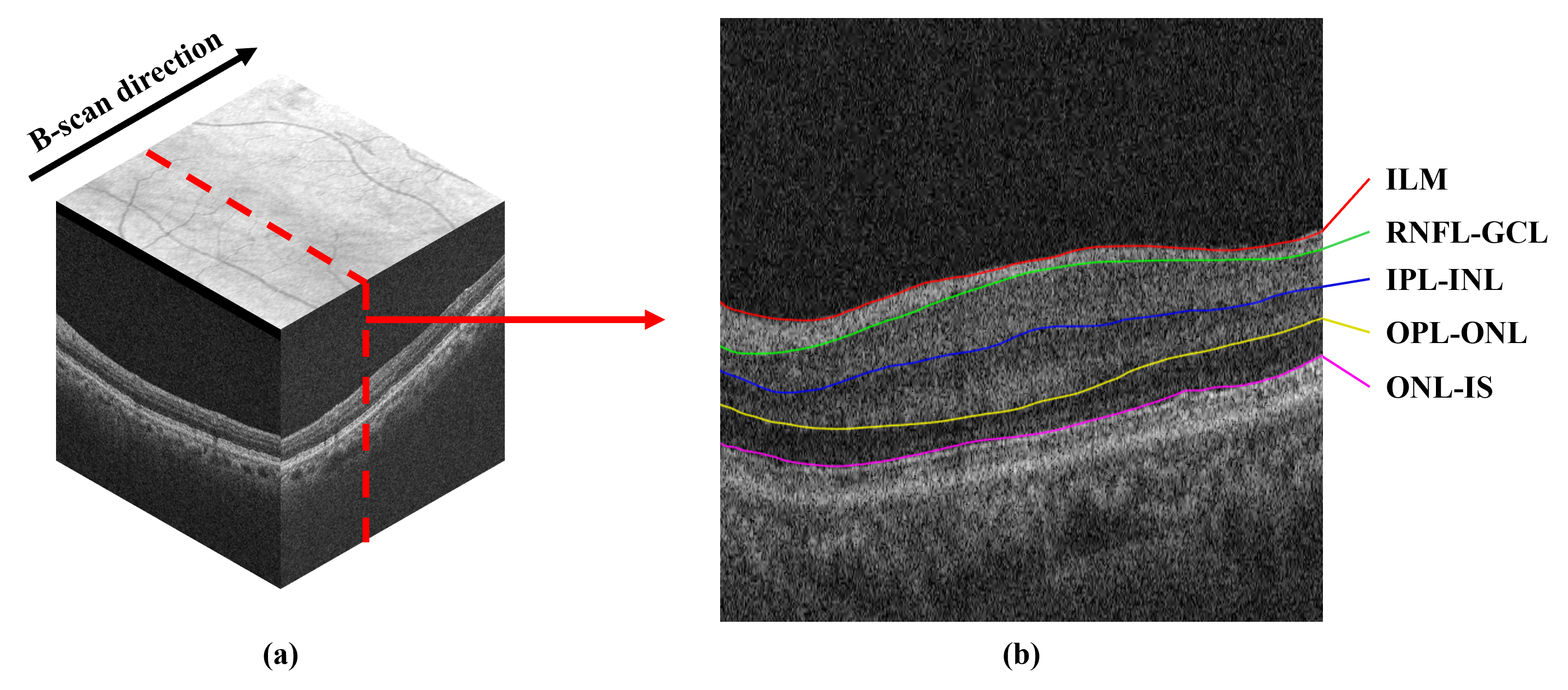}
    \caption{Example of our dataset. (a) Structure of the macular cube. (b) B-scan from the middle of the macular cube with 5 target retinal layer boundaries annotated by experienced clinicians.}
    \label{fig:our_dataset}
\end{figure}

\subsection{Dataset}\label{sec:dataset}
To train and evaluate our method, we employed two datasets, Our dataset, and DUKE DME \cite{chiu2015kernel} dataset.

\paragraph{Our dataset.}
We constructed a clinical OCT dataset comprising randomly selected patient scans acquired using Zeiss Cirrus HD-OCT 5000 OCT imaging device (Carl Zeiss Meditec, Inc., Dublin CA).
The Institutional Review Board of Gangnam Severance Hospital, Yonsei University Health System approved this study protocol (IRB No. 3-2025-0314).
All the research procedures adhered to the tenets of the Declaration of Helsinki.
Due to the retrospective nature of this study, which used de-identified OCT images, the requirement for written informed consent was waived.
A total of 32 macular cube scans were collected under standard diagnostic conditions.
Each scan covers a 6\(\times\)6\(\times\)6 mm³ retinal volume and consists of 128 B-scans with a resolution of 1024\(\times\)512 pixels.
Experienced clinicians annotated all images to obtain high-quality ground-truth labels for retinal layer boundaries.
Fig.~\ref{fig:our_dataset} shows an example of our dataset which include the structure of the macular cube, and an OCT image with 5 annotated retinal layer boundaries.
Each case was categorized by its clinical diagnosis. The dataset includes 6 normal cases, 20 glaucoma cases, including normal tension glaucoma, primary open angle glaucoma, and angle closure glaucoma, and 6 cases with other retinal diseases such as proliferative diabetic retinopathy and diabetic macular edema.
The training set included 20 cases (4 normal, 12 glaucoma, and 4 other disease cases), the validation set included 2 glaucoma cases, and the test set included 10 cases (2 normal, 6 glaucoma, and 2 other disease cases).
Table~\ref{tab:dataset_distribution} shows the distribution of diagnostic categories in each dataset.

\paragraph{DUKE DME dataset.}
For a general case evaluation, we used the DUKE diabetic macular edema (DME) dataset~\cite{chiu2015kernel}, which is a publicly available dataset for retinal layer segmentation.
It contains OCT images from patients with DME and provides expert-annotated ground truth.
The dataset consists of 610 B-scans from 10 patients, with a resolution of 496\(\times\)768 pixels.
Each patient scan consists of 61 B-scans, with annotations provided for 11 B-scans.
These annotated scans are centered at the fovea, including 5 frames on either side of the central foveal scan.
This dataset was only used for evaluation purposes and was not used for training.

\subsection{Comparative methods and evaluation metrics}
We conducted comparative experiments using several state-of-the-art retinal layer segmentation approaches, which consist of a traditional graph-based method~\cite{chiu2010automatic} and three deep learning-based methods: U-Net~\cite{ronneberger2015u}, ReLayNet~\cite{roy2017relaynet}, and FCRN~\cite{he2021structured}.
To ensure a fair comparison, we implemented all learning-based methods using the same architectural parameters, including network depth, number of channels, and convolutional kernel sizes, matching those of our proposed method.
Also, they were trained on the same training and validation datasets and evaluated on the same test dataset described in Section~\ref{sec:dataset} for consistent experimental conditions.
To ensure a consistent output representation, all methods were designed to output column-wise boundary positions.

As evaluation metrics, the differences between the predicted boundary positions and the ground-truth boundary positions were selected.
Specifically, we employed two metrics: Mean Absolute Distance (MAD) and Root Mean Square Error (RMSE).
They are mathematically defined as follows
\begin{align}
\text{MAD} &= \frac{1}{W}\sum_{u=1}^{W} \left| p_u - g_u \right|\\
\text{RMSE} &= \sqrt{\frac{1}{W}\sum_{u=1}^{W} \left( p_u - g_u \right)^2 }
\end{align}
where \(W\) is the width of the OCT image, \(p_u\) is the predicted boundary position for the \(u\)-th column, and \(g_u\) is the ground truth boundary position for the \(u\)-th column.
In both metrics, lower values indicate better performance.
To evaluate the accuracy of each boundary, the metrics were first computed separately for each layer.
In addition, the averages across all boundaries were calculated to evaluate the overall performance.

\subsection{Implementation details}
All OCT images from both our dataset and DUKE DME dataset were processed using identical pre-processing steps described in Section~\ref{sec:preprocessing}, and resized to 512\(\times\)512.
During training, we set an initial learning rate to 0.001 and a batch size to 4 and used the Adam optimizer with a weight decay of 0.001.
For the 2.5D segmentation approach, the number of consecutive slices was set to 3.
All learning-based methods, including the proposed method and comparative methods, were implemented in PyTorch, and trained on NVIDIA GeForce RTX 4080 SUPER GPU for 200 epochs.

\begin{table}[t]
\centering
\caption{Evaluation results on our dataset among multiple methods. Mean absolute distance (MAD) and root mean squared error (RMSE) of the boundary positions are measured in pixel units, with mean and standard deviation. Lower is better. The bolded values are the best performing values.}
\label{tab:our_dataset_results}
\resizebox{0.8\textwidth}{!}{%
\begin{tabular}{@{}lcccccc@{}}
\multicolumn{7}{c}{(a) MAD (Std)} \\
\toprule
Method & ILM & RNFL-GCL & IPL-INL & OPL-ONL & ONL-IS & Average \\
\midrule
Graph~\cite{chiu2010automatic} & 3.21 (1.57) & 9.06 (3.04) & 5.74 (2.77) & 4.69 (3.35) & 2.26 (0.86) & 4.99 (1.87) \\
Basic UNet~\cite{ronneberger2015u} & 2.43 (0.80) & 3.26 (1.62) & 2.84 (1.00) & 2.80 (0.54) & 2.03 (0.55) & 2.67 (0.80) \\
ReLayNet~\cite{roy2017relaynet} & 2.53 (0.91) & 3.09 (0.89) & 2.89 (0.97) & 2.85 (0.59) & 2.06 (0.63) & 2.69 (0.68) \\
FCRN~\cite{he2021structured} & 3.07 (0.90) & \textbf{2.66 (0.99)} & 2.86 (0.46) & 2.98 (0.45) & 2.05 (0.31) & 2.64 (0.58) \\
Ours & \textbf{2.36} (0.80) & 2.93 (0.67) & \textbf{2.56} (0.39) & \textbf{2.70} (0.47) & \textbf{1.85} (0.29) & \textbf{2.48} (0.48) \\
\bottomrule
\addlinespace[0.5em]
\multicolumn{7}{c}{(b) RMSE (Std)} \\
\toprule
Method & ILM & RNFL-GCL & IPL-INL & OPL-ONL & ONL-IS & Average \\
\midrule
Graph~\cite{chiu2010automatic} & 4.09 (1.92) & 13.85 (3.25) & 8.02 (3.68) & 6.70 (4.38) & 3.18 (1.35) & 7.17 (2.31) \\
Basic UNet~\cite{ronneberger2015u} & 3.66 (1.28) & 5.52 (3.89) & 4.71 (3.13) & 4.22 (1.88) & 2.97 (1.03) & 4.22 (2.12) \\
ReLayNet~\cite{roy2017relaynet} & 3.60 (1.41) & 4.97 (2.10) & 4.82 (3.33) & 4.40 (2.33) & 3.24 (1.63) & 4.20 (1.88) \\
FCRN~\cite{he2021structured} & 3.17 (1.11) & 4.32 (1.41) & 3.54 (0.71) & 3.66 (0.82) & 2.41 (0.40) & 3.42 (0.82) \\
Ours & \textbf{3.04} (1.02) & \textbf{3.99} (0.97) & \textbf{3.45} (0.62) & \textbf{3.61} (0.77) & \textbf{2.41} (0.36) & \textbf{3.30} (0.67) \\
\bottomrule
\end{tabular}
}
\end{table}

\begin{figure}[t]
    \centering
    \includegraphics[width=\linewidth]{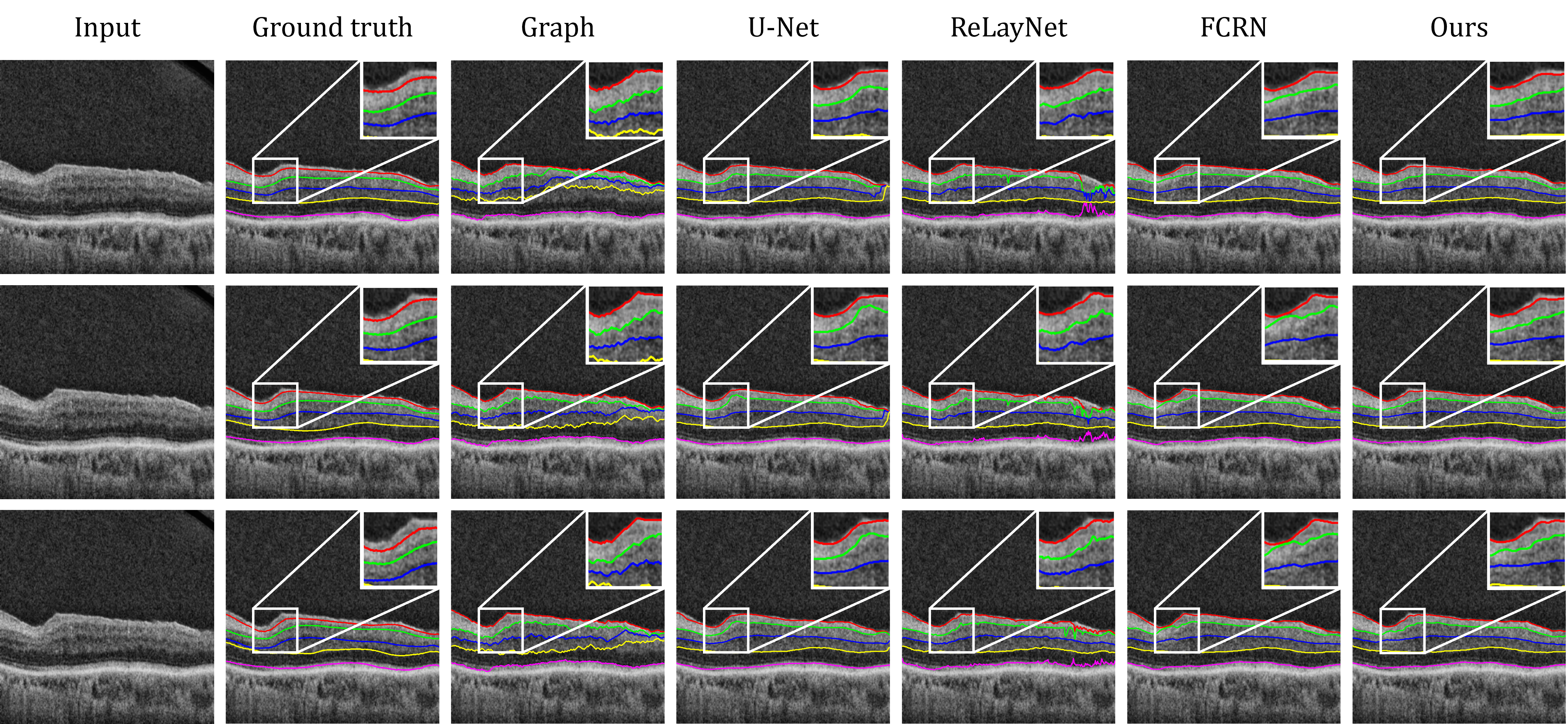}
    \caption{Qualitative results of the comparison methods on our dataset. The rows represent three consecutive B-scans. An enlarged region is shown in the upper-right corner of each image.}
    \label{fig:our_result_image}
\end{figure}

\subsection{Experimental results}

\subsubsection{Experiments on our dataset}
The proposed method and comparative methods were trained and tested on our dataset.
Table~\ref{tab:our_dataset_results} shows the quantitative results of our dataset.
The results include the mean and standard deviation of MAD and RMSE for each boundary across different methods.
On average, our method achieved the best performance with MAD of 2.48, and RMSE of 3.30, outperforming all comparative methods.
Notably, our method achieved not only the lowest mean MAD and RMSE but also the lowest standard deviation on average across boundaries for both metrics. 
This indicates that the predictions of our method are more stable and less affected by local outliers compared to other methods. 
Fig.~\ref{fig:our_result_image} shows qualitative results on three consecutive B-scans from our dataset.
While comparative methods show inconsistent boundary predictions at the same pixel locations across adjacent slices, the proposed method shows high cross-slice consistency.
In addition, the comparative methods produce irregular and jagged predictions in low contrast regions, whereas the proposed method produces smooth boundary estimates.
These results demonstrate that the proposed method effectively leverages contextual information, enabling consistent and robust boundary predictions even in low quality regions.

\begin{table}[t]
\centering
\caption{Evaluation results on the DUKE DME dataset among multiple methods. Mean absolute distance (MAD) and root mean squared error (RMSE) of the boundary positions are measured in pixel units, with mean and standard deviation. Lower is better. The bolded values are the best performing values.}
\label{tab:duke_dme_dataset_results}
\resizebox{0.8\textwidth}{!}{%
\begin{tabular}{@{}lcccccc@{}}
\multicolumn{7}{c}{(a) MAD (Std)} \\
\toprule
Method & ILM & RNFL-GCL & IPL-INL & OPL-ONL & ONL-IS & Average \\
\midrule
Graph~\cite{chiu2010automatic} & \textbf{1.86} (0.40) & 8.05 (2.96) & 13.37 (5.05) & 16.37 (6.52) & \textbf{3.05} (1.00) & 8.54 (3.07) \\
Basic UNet~\cite{ronneberger2015u} & 4.44 (4.07) & 5.66 (2.44) & 7.93 (3.28) & 12.83 (1.46) & 7.18 (1.46) & 7.61 (3.15) \\
ReLayNet~\cite{roy2017relaynet} & 3.90 (1.45) & 10.51 (4.99) & 10.24 (4.66) & 9.26 (3.38) & 14.98 (7.62) & 9.78 (4.01) \\
FCRN~\cite{he2021structured} & 3.70 (2.18) & 6.72 (2.81) & 5.30 (1.71) & 5.66 (1.68) & 4.56 (1.35) & 5.19 (1.76) \\
Ours & 2.96 (1.10) & \textbf{5.65} (1.25) & \textbf{5.04} (1.39) & \textbf{5.10} (1.78) & 4.67 (1.28) & \textbf{4.68} (1.11) \\
\bottomrule
\addlinespace[0.5em]
\multicolumn{7}{c}{(b) RMSE (Std)} \\
\toprule
Method & ILM & RNFL-GCL & IPL-INL & OPL-ONL & ONL-IS & Average \\
\midrule
Graph~\cite{chiu2010automatic} & \textbf{2.37} (0.50) & 12.81 (4.17) & 18.80 (5.11) & 22.48 (6.65) & \textbf{4.74} (1.24) & 12.24 (3.39) \\
Basic UNet~\cite{ronneberger2015u} & 10.02 (10.99) & 11.55 (7.18) & 15.18 (6.22) & 21.02 (12.15) & 8.77 (1.87) & 13.31 (6.34) \\
ReLayNet~\cite{roy2017relaynet} & 7.42 (3.47) & 18.62 (8.61) & 18.21 (8.10) & 15.80 (6.86) & 21.24 (10.06) & 16.26 (6.71)  \\
FCRN~\cite{he2021structured} & 7.58 (7.36) & 11.29 (6.52) & 9.19 (5.05) & 9.50 (3.42) & 6.04 (1.92) & 8.72 (4.69)  \\
Ours & 5.06 (3.14) & \textbf{8.30} (2.55) & \textbf{7.71} (2.72) & \textbf{8.98} (4.54) & 5.69 (1.44) & \textbf{7.15} (2.53)  \\
\bottomrule
\end{tabular}
}
\end{table}

\begin{figure}[t]
    \centering
    \includegraphics[width=0.8\linewidth]{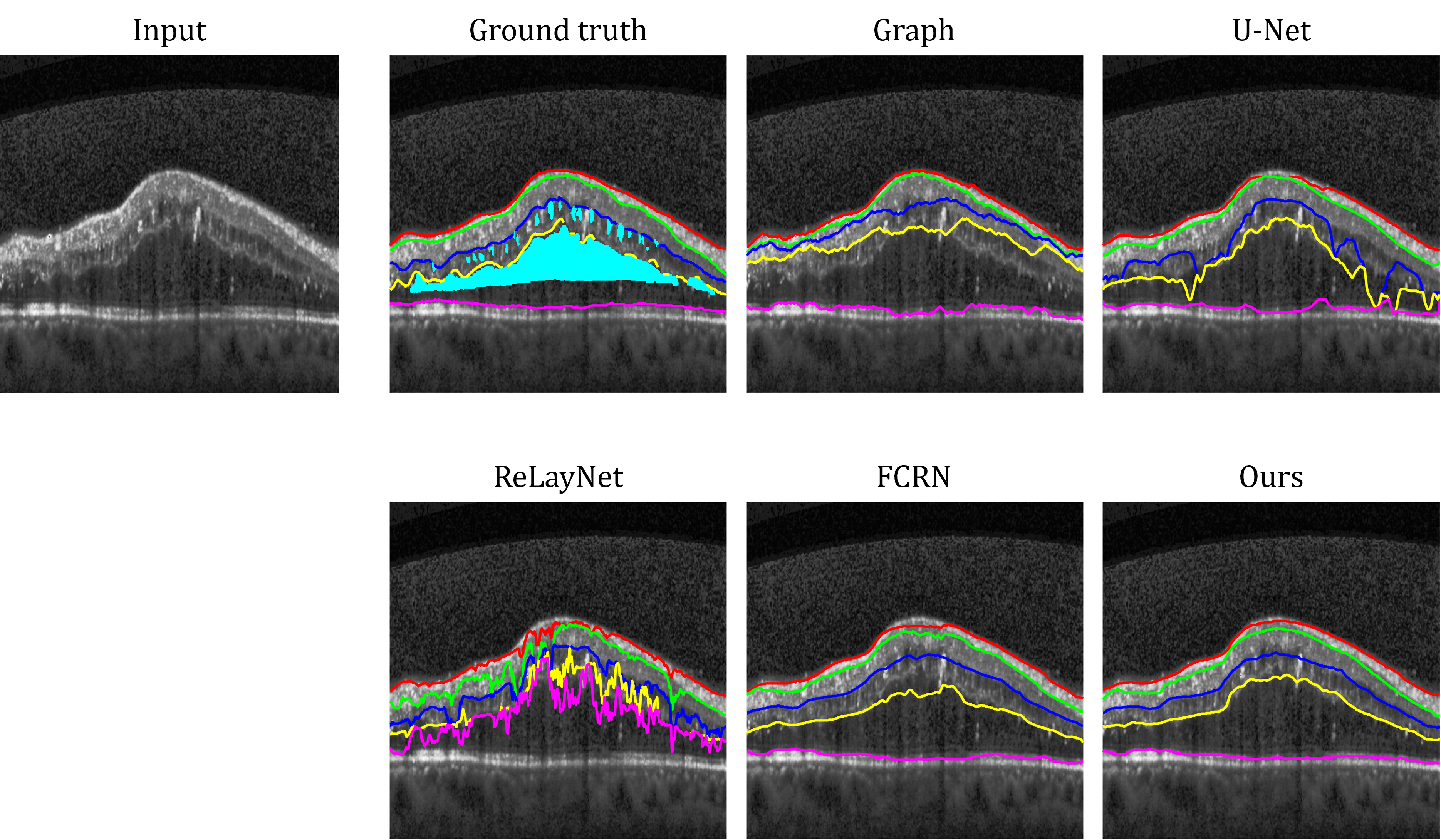}
    \caption{Qualitative results of the comparison methods on the DUKE DME dataset. The cyan regions in the ground truth indicate the fluid regions.}
    \label{fig:duke_dme_result_image}
\end{figure}

\subsubsection{Experiments on the DUKE DME dataset}
We evaluated the generality of our method by testing both our approach and the comparative methods on the DUKE DME dataset. 
All models were trained using only our training dataset, and no additional fine-tuning was performed. 
Table~\ref{tab:duke_dme_dataset_results} shows the quantitative results of the DUKE DME dataset.
The results include the mean and standard deviation of MAD and RMSE for each boundary across different methods.
Some B-scans in the DUKE DME dataset contain fluid regions where retinal layers are not labeled, resulting in discontinuous ground truth boundaries.
To ensure a fair comparison, such columns were excluded from the computation of MAD and RMSE.
On average, our method achieved the best performance with MAD of 4.68, and RMSE of 7.15, outperforming all comparative methods.
In addition, our method also showed the lowest average standard deviation across boundaries for both MAD and RMSE, indicating greater robustness to local outliers and structural irregularities.
Fig.~\ref{fig:duke_dme_result_image} shows an example of qualitative results on the DUKE DME dataset.
These results highlight not only the strong generalization ability of our approach, but also its robustness in handling challenging cases, demonstrating the effectiveness of the proposed cross-slice feature fusion module on independent datasets with different pathological characteristics.

\section{Conclusion}\label{sec:conclusion}



In this study, we proposed a deep-learning based 2.5D retinal layer segmentation framework for glaucoma assessment.
We developed a novel CFF module incorporated into a U-Net-like architecture to leverage contextual information.
We analyzed the results to verify that contextual information was effectively exploited and focused on accuracy and inter-slice consistency.
We assessed our method using a clinical dataset and a publicly available DUKE DME dataset.

The experimental results for our clinical dataset demonstrate that our method achieved the lowest average MAD and RMSE values.
Notably, our method achieved the lowest mean MAD and RMSE as well as the lowest standard deviation on average across the boundaries.
This indicates that the predictions of our model are more stable and less affected by local outliers than those of the other methods.
This robustness can be attributed to the incorporation of contextual information through the CFF module.
In the OCT images, severe speckle noise often degrades local image quality, making 2D segmentation approaches prone to spike-like errors when relying on a single slice.
Our method effectively suppresses noise-induced fluctuations by leveraging the adjacent slice information and produces more consistent boundary estimations, which are quantitatively evidenced by decreased standard deviations. Qualitative results revealed that our model maintained high consistency across consecutive slices.

Similar to the experimental findings on our dataset, the findings on the DUKE DME dataset showed that our approach achieved the lowest mean and standard deviation for MAD and RMSE, validating its accuracy and consistency.
Moreover, the results for the DUKE DME dataset demonstrate the generalizability of our method.
This dataset includes diverse cases, such as DME, which were not included in our training dataset, and the B-scans in this dataset were acquired from another OCT imaging device.
These results demonstrate the strong generalization ability of our approach and its robustness in handling challenging cases with varying pathological characteristics, validating the effectiveness of the proposed CFF module on independent datasets.
Such robustness is especially valuable from a clinical perspective.
Our method can potentially minimize the requirement for manual correction by clinicians by maintaining segmentation consistency even in noisy B scans or eyes with pathologies, thereby streamlining the diagnostic process for glaucoma.

Our study had some limitations that warrant further discussion.
First, the proposed framework was exclusively trained and validated using macular cube scans.
While detecting macular ganglion cell loss is important for identifying early stage glaucoma and monitoring advanced cases, evaluating the peripapillary RNFL from optic disc scans remains the gold standard for diagnosis.
Consequently, the performance of our 2.5D CFF module on optic disc scans, which present varying anatomical challenges, such as major vessel shadows, has not yet been verified.
Future studies should aim to extend our framework to optic disc datasets to ensure its comprehensive applicability in glaucoma evaluation.

In conclusion, the proposed 2.5D retinal layer segmentation framework with the CFF module provides an effective and computationally efficient solution for accurate and consistent retinal layer segmentation.
We address the limitations of 2D and 3D based methods while retaining their advantages.
These results suggest that the proposed approach provides a powerful and efficient solution for anatomically faithful retinal layer segmentation, which is crucial for the diagnosis, staging, and longitudinal monitoring of glaucoma.

\begin{backmatter}
\bmsection{Funding}
This work was supported by the Basic Science Research Program through the National
Research Foundation of Korea (NRF-RS-2025-00562090 \& RS-2024-00405287), the National Research Foundation of Korea(NRF) grant funded by the Korea government(MOE)
(NO.4199990314094, Mechanical Technology Global Leader Program for Society Development), and the Gangnam Severance Hospital and the Institute of Engineering Research, Yonsei University (D-2025-0002 \& D-2023-0012).

\bmsection{Acknowledgment}
We gratefully acknowledge the contributions of our colleagues, who prepared the ground truth labels used in the segmentation model.

\bmsection{Disclosures}
The authors declare no conflicts of interest.

\bmsection{Data availability}
Data underlying the results presented in this paper are not publicly available at this time, since the data may be restricted for privacy reasons.

\end{backmatter}

\bibliography{main}

\end{document}